\documentclass[10pt,twocolumn,letterpaper]{article}

\usepackage{times}
\usepackage{epsfig}
\usepackage{graphicx}
\usepackage{amsmath}
\usepackage{amssymb}

\usepackage[pagebackref=true,breaklinks=true,letterpaper=true,colorlinks,bookmarks=false]{hyperref}

\usepackage{enumitem}

\begin{document}

\title{Sampling Strategy for Fine-Tuning Segmentation Models to Crisis Area under Scarcity of Data}

\author{Adrianna Janik\\
Montreal Institute for Learning Algorithms
\and
Kris Sankaran\\
Montreal Institute for Learning Algorithms
}
\date{June 16th 2019}
\maketitle

\begin{abstract}
   The use of remote sensing in humanitarian crisis response missions is well-established and has proven relevant repeatedly. One of the problems is obtaining gold annotations as it is costly and time consuming which makes it almost impossible to fine-tune models to new regions affected by the crisis. Where time is critical, resources are limited and environment is constantly changing, models has to evolve and provide flexible ways to adapt to a new situation. The question that we want to answer is if prioritization of samples provide better results in fine-tuning vs other classical sampling methods under annotated data scarcity? We propose a method to guide data collection during fine-tuning, based on estimated model and sample properties, like predicted IOU score. We propose two formulas for calculating sample priority. Our approach blends techniques from interpretability, representation learning and active learning. We have applied our method to a deep learning model for semantic segmentation, U-Net, in a remote sensing application of building detection - one of the core use cases of remote sensing in humanitarian applications. Preliminary results shows utility in prioritization of samples for tuning semantic segmentation models under scarcity of data condition.

\end{abstract}

\section{Introduction}

Remote sensing has been extensively used by humanitarian aid teams, and collaborative mapping has become a crucial part of crisis response. Some of the success stories includes response after the Haiti and Nepal earthquakes, and Ebola outbreak in North West Africa in 2014 \cite{hunt_crowdsourced_2019}. Despite engagement of thousands of volunteers and NGOs there is still a place for improvement and making interventions such as \href{https://www.missingmaps.org/}{Missing Maps} more effective, therefore allowing Humanitarian Aid teams to be prepared better for handling the crisis. 

One of the problems that has emerged with the introduction of deep learning solutions in the field is their lack of integration with current humanitarian workflows. 
Although it is a natural idea -- the improvement of annotating new maps with the usage of deep learning models to automate the process -- such a system is not a part of typical workflows. One of the reasons is trust and reliability of predictions. A model that works on one geographic area might not work on another.

The aim of this paper is to present a method of selecting a subsample of a new dataset for manual labeling to be used during fine-tuning. The approach is to provide an estimated priority list of samples, based on estimated IOU scores. The goal is to assign higher importance to areas of interest, choosing instances that maximizes variety of instances. We can look at this problem in two ways: in the notion of improvement current model based on the knowledge of errors on the test set and in the notion of fine-tuning the model specifically to a new region. Improving the model over parts that poorly generalize on the data coming from the same distribution might be seen as less important then retraining the model to cover new areas, as we can assume that the deployed model had acceptable performance.

On which data shall we fine-tune it, if we cannot obtain labels for the whole region? How can we distribute the task of manual labeling to volunteers that will be the most effective and time efficient?

\subsection{Sampling Strategy Desiderata}
In the ideal scenario we would like to have a sampling method for selecting instances for annotation to be used in fine-tuning, that would: 
\begin{itemize}[noitemsep]
\item guarantee significantly better performance (over initial model)
\item use the least possible number of annotated samples to obtain it
\item prioritize areas of highest importance
\item not be undesirably biased
\end{itemize}



\subsection{How Interpretability Can Help?}

Model interpretability is inherently connected with the requirements of the individuals who are expected to thoroughly understand the problem and make a justified decision based on data. Lack of interpretability makes adoption of AI solutions hard, especially when the decision makers using deep learning model predictions are under enormous pressure. Why should they trust an oracle?
In the interpretability literature, most attention is focused on understanding black-box models \cite{guidotti_survey_2018} in classification tasks \cite{ribeiro_why_2016,kim_interpretability_2017}, but many problems ranging from medicine through agriculture and crisis response in humanitarian aid are tackled by semantic segmentation models. The absence of interpretability for these canonical problems in computer vision is concerning. Overall performance measures of black-box models may not be enough to justify the use of model predictions in the field.

\begin{figure}
    \centering
    \includegraphics[width=0.4\textwidth]{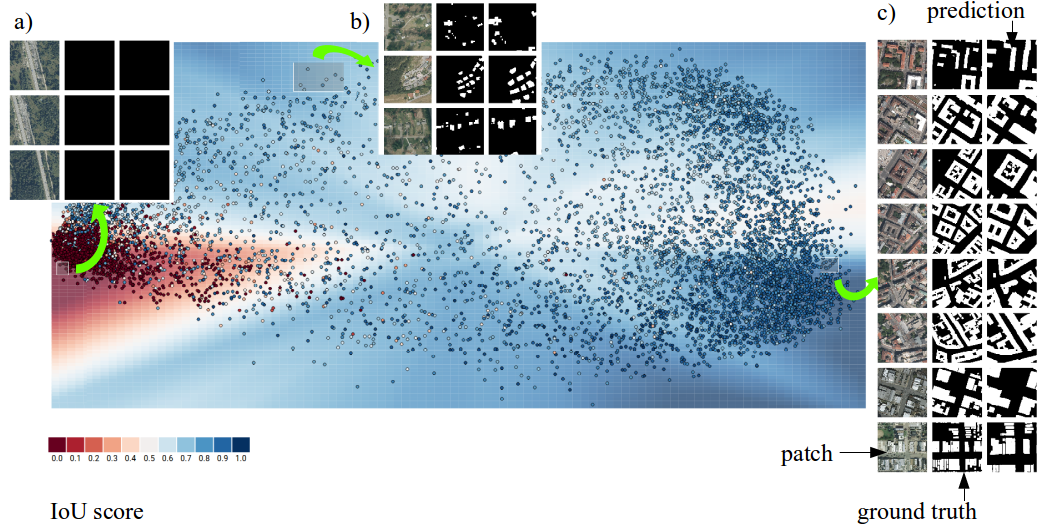}
    \caption{Visualization present samples from three qualitatively distinguishable regions a) that does not contain any buildings b) that contains few buildings c) highly urbanized with many buildings and with higher IoU score. We can see a bipolar nature of learned representation: undeveloped area and urbanized.}
    \label{fig:unet-vis}
\end{figure}

Understanding how a deep model works under-the-hood is crucial to our method.
Interpretability in image segmentation can be achieved by analyzing the latent representation in encoder-decoder network architectures, and we propose to use the same representation to guide the fine-tuning process. We focus on selection of generalization regions in building detection task. In Figure \ref{fig:unet-vis}, a visualization of U-Net latent space is presented. Each point is a representation of an input image in latent space, reduced with PCA. Associating the ground-truth mask with prediction from the model and the original image allows users to reason about the model. The color of points represents the IoU score. From the figure we can see that the latent representation has a somehow bipolar structure, on the left are patches that do not contain any buildings and on the right are highly urbanized one. This inspires us to seek a way to use this observation in a fine-tuning scenario.

\subsection{Mapping for Humanitarian Aid}

Crisis responders operate under challenging constraints, and AI solutions may offer information that allows them leverage their full potential and provide best crisis relief possible. To give a sense of how collaborative mapping works in humanitarian workflow, take an example of the application \href{https://mapswipe.org/}{MapSwipe}, a part of the \href{https://www.openstreetmap.org}{OpenStreetMap} ecosystem. Currently, volunteers have to manually choose map tiles that contain images of undeveloped areas, and exclude them from further processing. Filtering down to those that have desired features present is tedious, especially if images have the same geography. After the region affected by the crisis has been processed through MapSwipe, the narrowed down region goes to the \href{https://tasks.hotosm.org/}{Humanitarian OpenStreetMap Tasking Manager}, which assigns priorities to submitted regions and connects volunteers with a task for mapping (annotating map tiles with, for example, buildings). This part is done manually. In a building detection task, we could use a smarter way of prescreening tiles with our method that can instantly prioritize regions with higher probability of being inhabited. An example can be seen in Figure \ref{fig:unet-vis}.

\subsection{Why Prioritization?}

Why there is a need for prioritization of samples? Re-training a model on the entire available data for a new geographic region is not possible simply because there are no data that are annotated at this point. Gold annotations have to be done by people, and this directly translates to much longer time and cost for obtaining them. When the crisis occurs and an area needs to be remapped or mapped for the first time, people lives are at stake. On the other hand volunteers are contributing their own free time to help, so we need to make sure that this time is used to its highest potential value.

We considered several sample prioritization factors, which will be described in the following sections: 
\begin{itemize}[noitemsep]
    \item outliers
    \item clusters of errors
    \item orphaned or weak clusters (which we defined as clusters that are not found in training dataset)
    \item strong clusters
\end{itemize}

Choosing instances to label is one of the problem addressed by active learning. One interesting user-centric method, 
\cite{bernard_towards_2018}, inspired us to explore the user perspective of semantic segmentation. We emphasize that this increases the interpretability of our method.

                  

\section{Method}

Our method was designed for fine-tuning the U-Net semantic segmentation model \cite{ronneberger_u-net:_2015}, though in the future we plan to explore other networks with encoder-decoder architectures. It requires access to a trained U-Net segmentation model, a training dataset, and activations at the bottleneck layer. In experiments, we trained our own network, but this section (except \ref{unet}) only on fine-tuning. To evaluate segmentation predictions with respect to ground truth we used Intersection over Union score, defined as $$IoU(y,\hat{y}) = \frac{overlapping \; area}{union \; area}$$

\subsection{U-Net}\label{unet}
U-Net is a deep network commonly used in segmentation problems. It learns a reduced representation of an image through a down-sampling path, while at the same time preserving localized information about desired properties through an up-sampling path, which is used to make a prediction. 

Each component is composed of convolutional layers going  down and transposed convolutions going up, with max-pooling layers in between. Down-sampling is responsible for reducing the input image to a concise representation, while up-sampling retrieves localized information for the network's output. The latent representation referred in this work is represented by activations of the bottleneck layer - the layer that contains the quintessence of analyzed the image. 

\subsection{Algorithm}

We refer to training data that was used to train original model as core-training data and for data used in fine-tuning fine-tuning data.

We use the following strategy for prioritizing samples for fine-tuning the model to generalize to the new area:
\begin{enumerate}[noitemsep]
\item Collection of activations for core-training and fine-tuning datasets
\item  Dimensionality reduction of all collected activations (e.g. PCA)
\item  IOU prediction based on core-training data for fine-tuning data
\item  Clustering samples by their latent representation in two phases:
       a) clustering only core-training data with extra dimension of IOU
       b) classification of fine-tuning data into detected clusters
\item calculating priority score \ref{ps}
\item presenting subset of samples from highest to lowest priority score to the model
\end{enumerate}

\subsection{Priority Score}\label{ps}

Prioritization is based on priority score which two versions, \ref{v1} and \ref{v2}, proposed below.

In experiments, we only explored a basic priority score formula \ref{v1}, but it has limitations that we try to overcome in \ref{v2}.

\subsubsection{Basic Priority Score}\label{v1}

This score is defined as,
$$
    BPS = a*dist + b*(1-IoU)
$$
where: \\
$a = 0.75$, $b = 0.25$ \\
$dist$ - distance from the closest cluster normalized to [0-1]\\
$IoU$ - predicted intersection of the union \\
$a,b$ - arbitrarily selected coefficients of importance of each component.

This formula gives a score between 0 and 1 that prioritizes samples that are far from clusters centroids and at the same time have have a low $IoU$ score. This formula aims at spotting new clusters that were not covered during training, but does not deal with the outliers in a desirable way. 

\subsubsection{Multiparty Priority Score}\label{v2}

Another proposition accounts for more features that may influence the importance of a given sample,
$$
    MPS = (a*orph + b*err + c*dist + d*(1-IoU))(1-LoOP)
$$
where: \\
$a = 0.5$, $b = 0.25$, $c = 0.2$, $d = 0.05$ \\
$orph$ - size of orphaned cluster that sample belongs to (if not then $orph = 0$)\\
$err$ - size of error cluster that sample belongs to (if not then $err = 0$)\\
$LoOP$ - Local Outlier Probability \cite{kriegel_loop:_2009} \\
$a,b,c,d$ - arbitrarily selected coefficients of importance of each component.

This formula decreases the probability of selecting outliers, thanks to $LoOP$ score, and also considers two other aspects that are not covered in the \ref{v1}: belonging to the error cluster, defined by clustering core-training data below $IoU$ level of 0.5 and classifying fine-tuning data into those clusters. Another new component is orphaned cluster which we define as cluster that was not detected among core-training data.

\section{Experiments}
To evaluate these proposals we ran set of experiments on Inria Aerial Labelling Dataset \ref{dataset}. Our first step was to split it into two separate datasets: core-training\ref{core-training-ds} and fine-tuning\ref{fine-tuning-ds}, each with standard inner split into training/validation/testing.

\subsection{Dataset}\label{dataset}
The Inria Aerial Labelling Dataset \cite{maggiori_can_2017} is an example of a well-explored labeled dataset for satellite imagery. The training set contains 180 color image tiles of size $5000 \times 5000$, covering a surface of $1500m \times 1500m$ each (at a 30cm resolution). There are 36 tiles for each region. It covers 5 regions Austin, Chicago, Kitsap County, Western Tyrol, Vienna. For the test set there were another 5 regions chosen: Bellingham, WA; Bloomington, IN; Innsbruck; San Francisco; Eastern Tyrol. It provides all together coverage of 810 km$^2$. Images were sliced into patches of size $572 \times 572$.

\subsubsection{Core-Training Dataset}\label{core-training-ds}
The core-training dataset  was composed of four out of all five cities in the original dataset: Austin, Chicago, Western Tyrol and Vienna.

\subsubsection{Fine-Tuning Dataset}\label{fine-tuning-ds}
The last city - Kitsap County was held out for fine-tuning dataset. 
The choice of Kitsap was motivated by selecting the worse building detection performance in the model presented in Inria Dataset paper \cite{maggiori_can_2017}, as this region seemed to be the most challenging for model generalization. 

\subsection{Segmentation Model}

\begin{figure}
    \centering
    \includegraphics[width=0.5\textwidth]{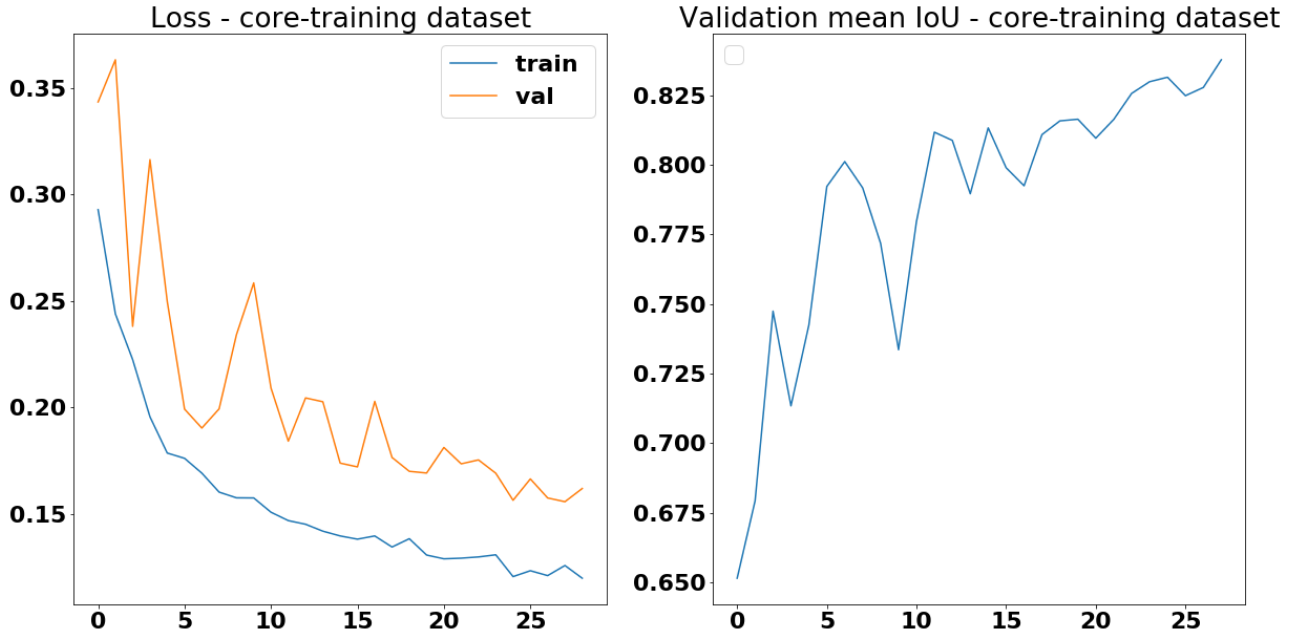}
    \caption{Training history of initial base model, on the left training and validation loss and on the right mean validation IoU score, at the end of training, the IoU was 83.8\%.}
    \label{fig:core-results}
\end{figure}

We trained U-Net network with Adam algorithm with batch normalization on the core-training dataset for 30 epochs. Training results can be seen in Figure \ref{fig:core-results}; the model scored 83.8\% IoU on validation set.

\subsubsection{Fine-Tuning Strategy}

To see how number and choice of instances influence fine-tuning, we ran set of experiments. In the fine-tuning dataset\ref{fine-tuning-ds} we had approximately 2200 patches (tiles of map) that represents the whole new region that we wanted to retrain our model to. From all the patches we created two series of small fine-tuning datasets with increasing number of labeled instances, one with instances selected by random sampling and the other with priorities-sampling. The two sequences of datasets each have between 250 and 2150 instances, with increment size of 100.

\begin{figure}
    \centering
    \includegraphics[width=0.5\textwidth]{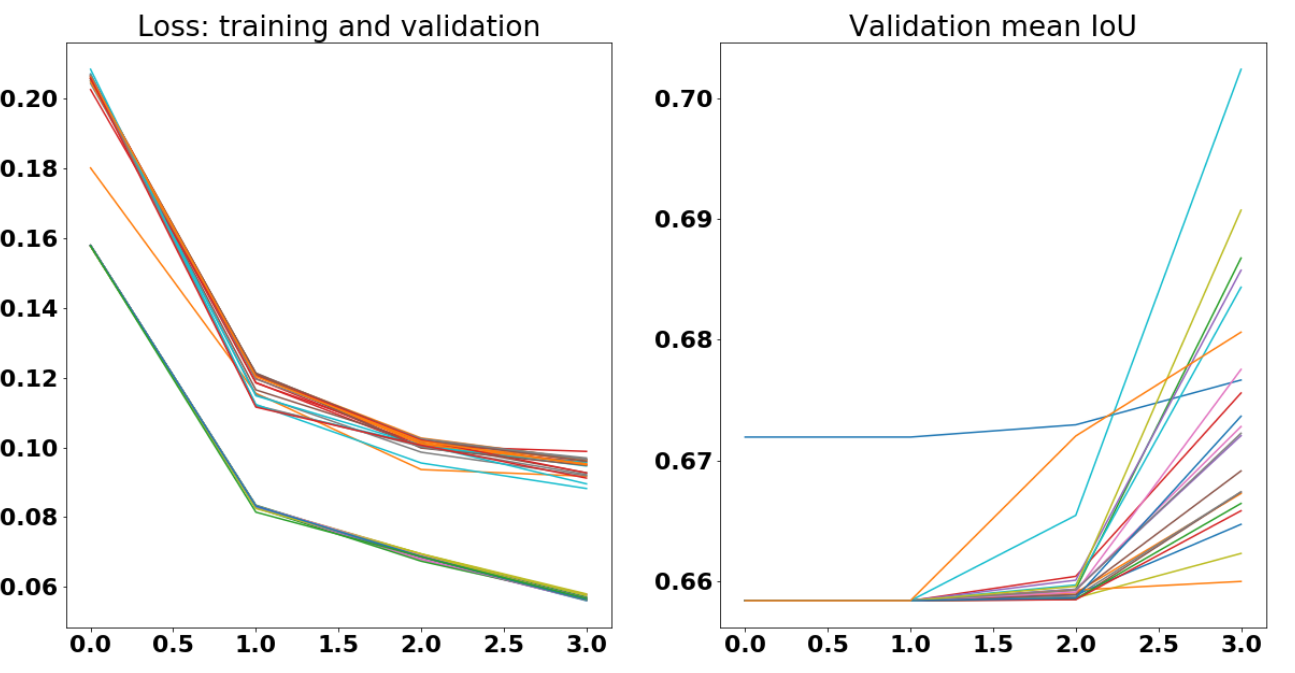}
    \caption{Fine-tuning history for sampling with prioritization. We can see that with bigger size of fine-tuning dataset increases average IoU score. For the sake of clarity legends were suppressed, upper set of plots represent validation loss and lower training loss.Each line represents one fine-tuning with subset of data.}
    \label{fig:priorities}
\end{figure}

\begin{figure}
    \centering
    \includegraphics[width=0.5\textwidth]{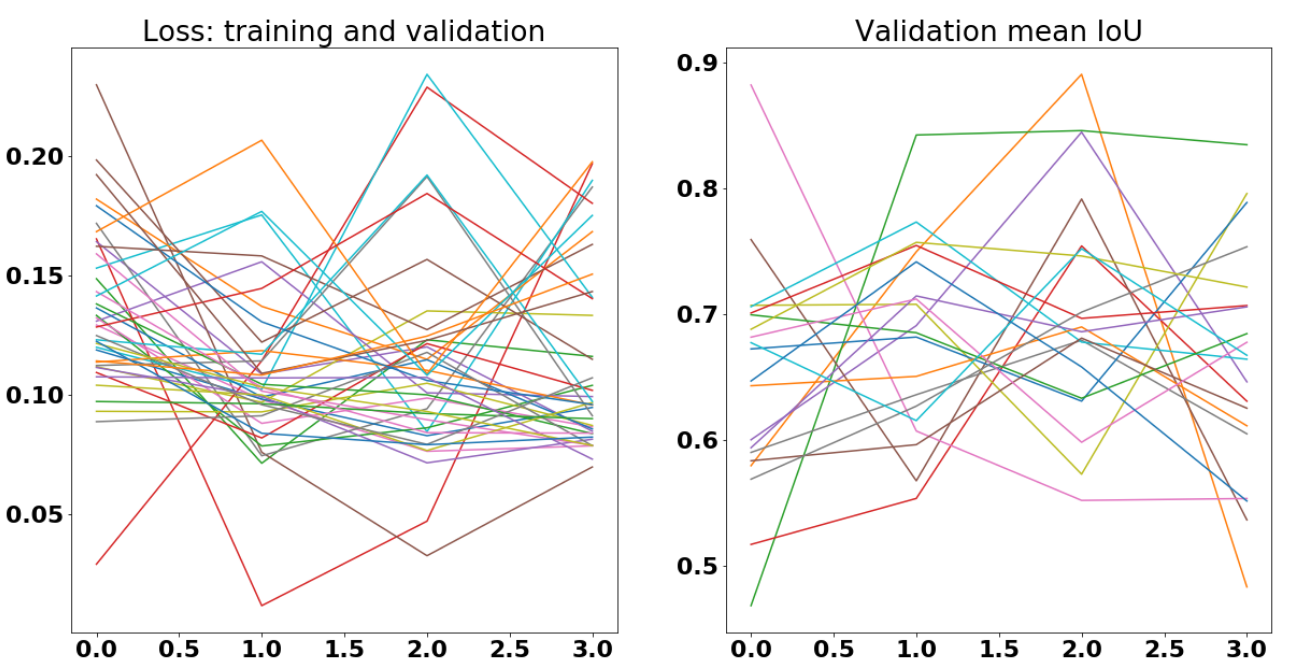}
    \caption{Fine-tuning history for random sampling. We cannot see any distinguishable pattern in fine-tuning history in this scenario. Also final IoU score is not consistent. For the sake of clarity legends were suppressed. Each line represents one fine-tuning with subset of data.}
    \label{fig:random}
\end{figure}

For each size of the fine-tuning dataset we fine-tune two models to compare how choice of training samples affects fine-tuning. The results of fine-tuning for random sampling is presented in Figure \ref{fig:random} and for sampling with priorities in Figure \ref{fig:priorities}.

\subsection{Results}

\begin{figure}
    \centering
    \includegraphics[width=0.5\textwidth]{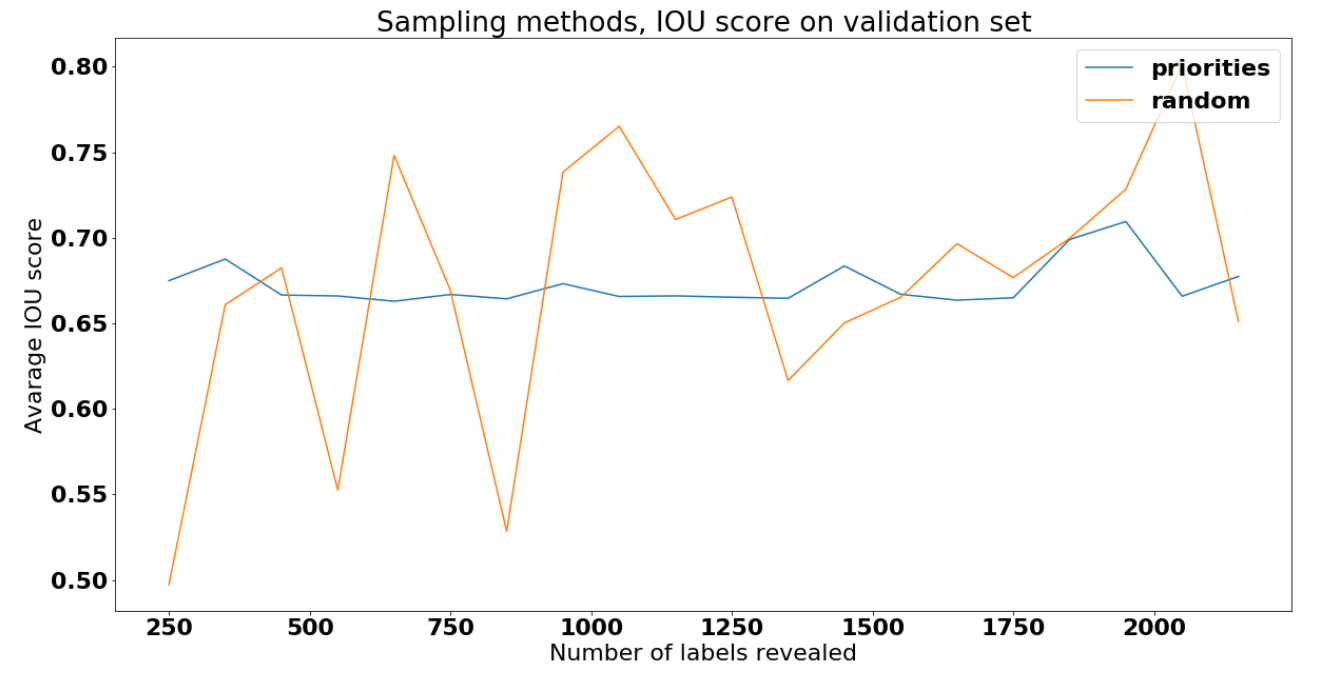}
    \caption{This chart presents results of 19 experiments of fine-tuning with different size of fine-tuning dataset. In each experiment two models were trained, one with samples selected randomly (orange line) and the other with samples selected based on priority score (blue one). With increasing size of fine-tuning random sampling tends to perform better, but under data scarcity condition prioritization becomes important.}
    \label{fig:comparison}
\end{figure}

\begin{figure}
    \centering
    \includegraphics[width=0.5\textwidth]{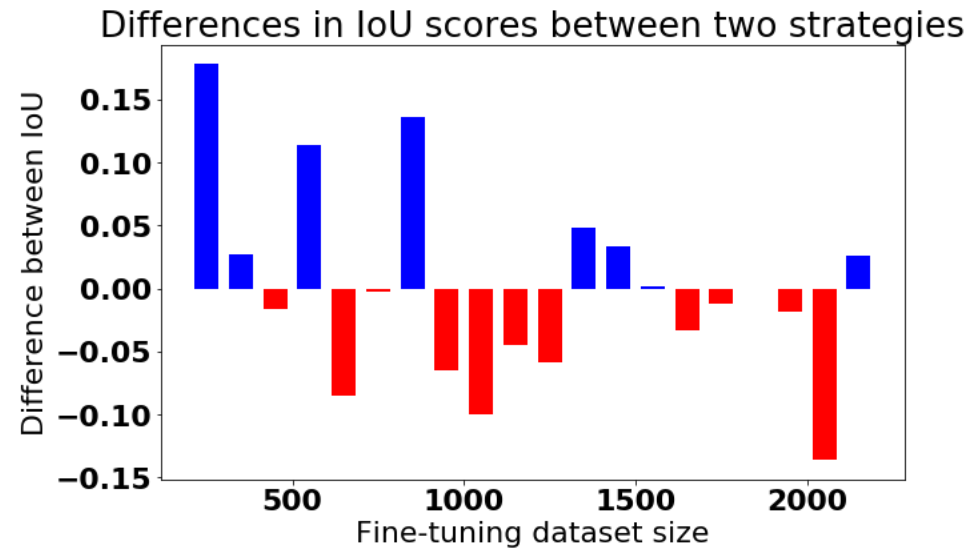}
    \caption{Bar chart with differences between IoU of two sampling strategies for all sizes of fine-tuning dataset. Blue bars indicates that the score was higher for prioritization strategy and red bar for random strategy. We can see that when the more data is available the better is model trained with randomly selected samples but the less samples are available fine-tuning with samples priorities gives better results.}
    \label{fig:diff_whole}
\end{figure}

\begin{figure}
    \centering
    \includegraphics[width=0.5\textwidth]{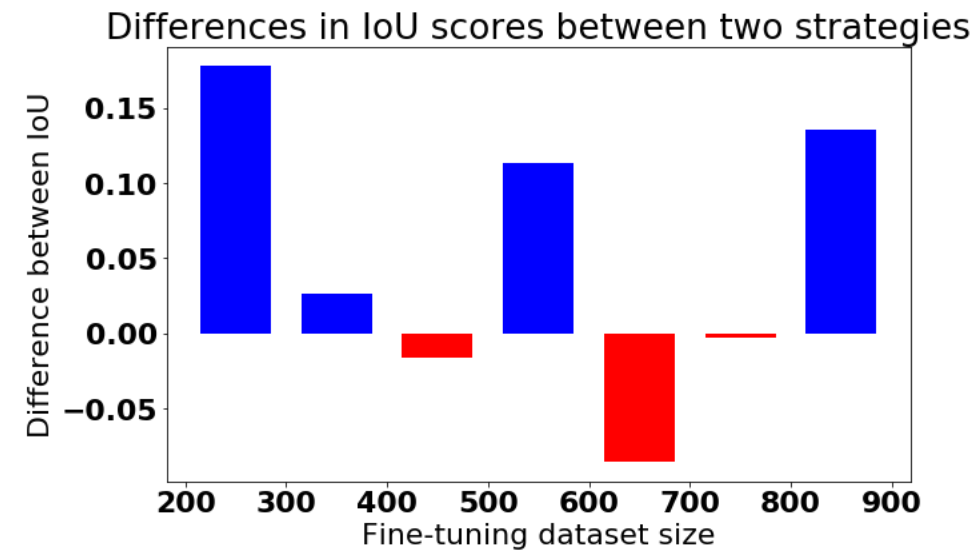}
    \caption{Bar chart with differences between IoU of two sampling strategies for initial seven sizes of fine-tuning dataset: that consecutively contains 250, 350, 450, 550, 650, 750, 850 samples. Blue bars indicates that the score was higher for prioritization strategy and red bar for random strategy. We can see that at least initial 20\% of samples might have a huge role in fine-tuning.}
    \label{fig:diff_iou}
\end{figure}

Experiments are summarized in Figure \ref{fig:comparison}. This figure shows results for the whole fine-tuning dataset, and we can see that when the number of annotated labels increases to the size of the whole fine-tuning dataset, random samples selection gives better results than any imposed order. This can be better seen in Figure \ref{fig:diff_whole}. The most important part however is captured between fine-tuning with only 10-40\% of data, as the goal is to provide better performance with less data (this would not matter if we had all the data annotated). This is shown in Figure \ref{fig:diff_iou}, where the initial 7 experiments are presented. The bar chart shows the differences in average $IoU$ scores between the two experiments per fine-tuning dataset size. In blue are bars where higher score was obtained by prioritization, and in red are those where a higher score was obtained by random selection of instances. This preliminary result shows that the first 500 samples in this scenario makes a difference. Another observation is that compared to random sample selection, prioritization seems to be more stable in final performance. With selecting samples at random with small fine-tuning datasets the results are highly variable.

\section{Conclusion}

Our main conclusion is that selection of samples for fine-tuning does matter if the size of fine-tuning dataset is highly limited. With random samples, we can get lucky or not, which should not be the basis for crisis response decisions. If deep models for building detection are to be included in the toolset of humanitarian mission teams in the field, they should be adaptable, flexible, and allow users to understand decisions.  The role of building detection task in humanitarian response is undeniable -- information about detected buildings is being used, for example, to estimate region populations. This knowledge guides humanitarian efforts in distribution of food, water and other basic resources for people affected by the crisis, and for creating strategies for epidemiology prevention.

The search for a method that minimizes the number of required labeled instances for fine-tuning, based on a smart prediction of their importance may therefore have a big impact on adoption of segmentation models into current humanitarian workflows.

{\small
\bibliographystyle{acm}
\bibliography{bibliography}
}

\end{document}